\documentclass[journal]{IEEEtran}

\ifCLASSINFOpdf
\else
\fi

\usepackage{graphicx}
\usepackage{amsmath}
\usepackage{multirow}
\usepackage{amsfonts}
\usepackage{url}
\usepackage{xcolor}
\usepackage{subcaption}
\usepackage[justification=centering]{caption}
\usepackage{float}
\DeclareMathOperator*{\argmin}{argmin} 

\begin{document}
\title{Deep Learning-Based Autonomous Driving Systems: A Survey of Attacks and Defenses}
\author{Yao~Deng,~\IEEEmembership{Student Member,~IEEE,}
        Tiehua~Zhang,~\IEEEmembership{Student Member,~IEEE,}
        Guannan~Lou,~\IEEEmembership{Student Member,~IEEE,}
        Xi~Zheng,~\IEEEmembership{Member,~IEEE,}
        Jiong~Jin,~\IEEEmembership{Member,~IEEE,}
        and~Qing-Long~Han,~\IEEEmembership{Fellow,~IEEE}
\thanks{This work was supported in part by the Australian Research Council Linkage Project under Grant LP190100676.} 
\thanks{Yao~Deng, Guannan~Lou and Xi~Zheng are with the Department of Computing,
Macquarie University, Sydney, NSW, 2109 Australia
(e-mail: yao.deng@hdr.mq.edu.au; lougnroy@gmail.com; james.zheng@mq.edu.au).} 
\thanks{Tiehua~Zhang, Jiong~Jin and Qing-Long~Han are with the School of Software and Electrical Engineering,
Swinburne University of Technology, Melbourne, VIC, 3122 Australia
(e-mail: tiehuazhang@swin.edu.au; jiongjin@swin.edu.au; qhan@swin.edu.au).}
}


\maketitle

\begin{abstract}
The rapid development of artificial intelligence, especially deep learning technology, has advanced autonomous driving systems (ADSs) by providing precise control decisions to counterpart almost any driving event, spanning from anti-fatigue safe driving to intelligent route planning. However, ADSs are still plagued by increasing threats from different attacks, which could be categorized into physical attacks, cyberattacks and learning-based adversarial attacks. Inevitably, the safety and security of deep learning-based autonomous driving are severely challenged by these attacks, from which the countermeasures should be analyzed and studied comprehensively to mitigate all potential risks. This survey provides a thorough analysis of different attacks that may jeopardize ADSs, as well as the corresponding state-of-the-art defense mechanisms. The analysis is unrolled by taking an in-depth overview of each step in the ADS workflow, covering adversarial attacks for various deep learning models and attacks in both physical and cyber context. Furthermore, some promising research directions are suggested in order to improve deep learning-based autonomous driving safety, including model robustness training, model testing and verification, and anomaly detection based on cloud/edge servers.
\end{abstract}


\IEEEpeerreviewmaketitle

\section{Introduction}
\label{Sec:introduction}
\IEEEPARstart{W}{ith} the development of artificial intelligence technologies, autonomous driving has been receiving considerable attention in both academia and industry. From 1987 to 1995,
the Eureka PROMETHEUS Project (PROgraMme for a European Traffic of Highest Efficiency and Unprecedented Safety) \cite{Eruka}, one of the earliest autonomous driving projects, was carried out by Daimler-Benz. In 2005, a famous autonomous driving competition called DARPA~\cite{DARPA} Grand Challenge was organized. Since then, numerous development/refinement on advanced autonomous driving systems (ADSs) have been proposed. For now, autonomous vehicles are still going through the transformation through five levels, from level 0 (no automation) to level 4 (high self-driving automation). Most of companies like Tesla \cite{telsaS} focus on the development of level 3 ADSs that could achieve limited self-driving in some conditions (e.g., on highway). The top runner Google Waymo \cite{googleS} is currently committed to research and industrializing on Level 4 ADSs that do not require human interaction in most circumstances. More importantly, a consensus has been reached that the advent of autonomous vehicles will improve people's driving experience significantly. However, research on self-driving vehicles is still in its infancy stage. Some critical issues, especially for issues related to safety, need to be well tackled before proceeding to the full-scale of industrialization. For instance, the recent Uber's vehicle's fatal accident \cite{uberkill} reveals the importance of prioritizing the research on the safety of autonomous driving.

Deep learning, the most popular technique of artificial intelligence, is widely applied in autonomous vehicles to fulfill different perception tasks as well as making real-time decisions. Figure~\ref{Fig:autonomous driving system} demonstrates the workflow and architecture of a deep learning-based ADS. In a nutshell, raw data collected by diverse sensors and high-definition (HD) map information from the cloud are first fed into deep learning models in the perception layer to extract the ambient information of the environment, after which different designated deep/reinforcement learning models in the decision layer kicks off the real-time decisions making process. For example, in Baidu Apollo~\cite{apollo}, which is the ADS applied in Baidu Go Robotaxi service~\cite{robotaxi}, several deep learning models are used in perception and decision modules. Tesla also deploys advanced AI models for object detection to implement Autopilot~\cite{autopilot}. However, there exist a number of issues against the further development of deep learning-based ADSs adopting this pipeline structure. First of all, sensors are vulnerable to numerous physical attacks, under which most of the sensors are no longer able to function as normal to collect data in good quality, or they may be adversely instructed to collect fake data, leading to a severe degradation of performance of all learning-based models in the following layers. Furthermore, recent research shows that deep neural networks are vulnerable to \textit{adversarial attacks} \cite{szegedy2013intriguing} that are designed specifically to induce learning-based models to wrong predictions. The most common adversarial attack is by constructing the so-called \textit{adversarial examples} that only have slight difference from the original inputs to baffle the neurons in the model. There are some results available from prior research literature that focus on investigating such adversarial attacks \cite{szegedy2013intriguing, goodfellow2014explaining, kurakin2016adversarial, kurakin2016adversarialphysical, tramer2017ensemble, carlini2017towards, chen2018ead, moosavi2016deepfool, su2019one, moosavi2017universal, poursaeed2018generative, xiao2018generating, liu2019perceptual}, exhibiting the level of significance of these threats to the safety of deep learning-based ADSs. 

 The potential risks of ADSs have the effects on the development and deployment of autonomous vehicles in industry. If autonomous vehicles cannot ensure safety when they are running, they will not be accepted by the public. Therefore, it is essential to figure out whether deep learning-based ADSs are vulnerable, how they could be attacked, how much damage can be caused by attacks, and what measurements have been proposed to defend these attacks. The industry needs this information and further insights to improve their development of safety and robustness of ADSs.   
 Though safety threats and defenses of autonomous vehicles and autonomous vehicular networks have been studied before~References \cite{ren2019security, survey2020},
 none of these investigated on 
 security problems in deep learning-based ADSs. On the other hand, most of researchers on safety deep learning focus on adversarial attacks on the image classification task. For example, in~\cite{akhtar2018threat} and~\cite{yuan2019adversarial}, adversarial attacks and defenses for computer vision tasks were thoroughly introduced. However, related analysis on attacks and defenses on deep learning systems for more complicated autonomous driving tasks were not covered in these works.

Therefore, in this paper, we conduct a comprehensive survey that pulls together the recent research efforts on the workflow of deep learning-based ADSs, the state-of-the-art attacks and the corresponding defending strategies. The contributions of this paper are listed as follows:
\begin{itemize}
\item A variety of attacks towards the pipeline of deep learning-based ADSs are reviewed and analyzed in detail.  
\item The state-of-the-art attacks and the defending methods in deep learning-based ADSs are comprehensively elucidated.
\item Future research directions of applying new attacks as well as securing and improving the robustness of deep learning-based ADSs are proposed.
\end{itemize}

The paper is organized as follows: Section \ref{Sec:workflow} introduces the detail of pipeline in deep learning-based ADSs and possible threat models adopted by adversaries against the systems. Section \ref{Sec:threat model} walks through 
different attacks that could occur in the pipeline as well as their threat models. Section \ref{defense} summarizes defenses corresponding to the mentioned attacks and discusses their effectiveness in protecting ADSs. Section~\ref{Sec:future direction} reveals future research directions for securing ADSs. Section~\ref{Sec: conclusion} draws the conclusion.

\section{Workflow of deep learning-based ADSs}
\label{Sec:workflow}
\begin{figure}[tb!]
\centering
\includegraphics[width = .4\textwidth]{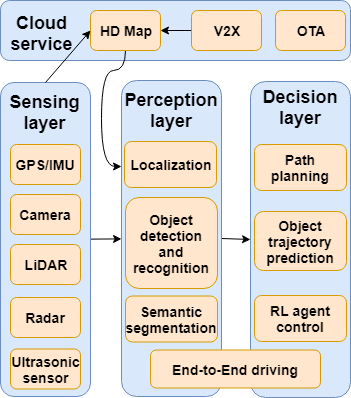}

\caption{ADS architecture}
\label{Fig:autonomous driving system}
\end{figure}

A deep learning-based ADS is normally composed of three functional layers, including a sensing layer, a perception layer and a decision layer, as well as an additional cloud service layer as shown in Figure \ref{Fig:autonomous driving system}. In the sensing layer, heterogeneous sensors such as GPS, camera, LiDAR, radar, and ultrasonic sensors are used to collect real-time ambient information including the current position and spatial-temporal data (e.g. time series image frames). The perception layer, on the other hand, contains deep learning models to analyze the data collected by the sensing layer and then extract useful environmental information from the raw data for further process. The decision layer would act as a decision-making unit to output instructions concerning the change of speed and steering angle based on the extracted information from the perception layer. The following part of this section will unveil the workflow of a deep learning-based ADS.

\subsection{The sensing layer}
The sensing layer encompasses heterogeneous sensors to collect surrounding information around an autonomous vehicle. The most preferred sensors adopted and deployed by leading autonomous driving vehicle companies like Baidu are GPS/Inertial Measurement Units (IMU), cameras, Light Detection and Ranging (LiDAR), Radio Detection and Ranging (Radar), and ultrasonic sensors. More specifically, GPS could provide absolute position data through the help of geostationary satellites, while IMU provides orientation, velocity and acceleration data. 

Cameras are also used to capture visual information around an autonomous vehicle, providing abundant information to the perception layer to analyze so that the vehicle could recognize traffic signs and obstacles. Furthermore, LiDAR is also applied to help detect objects by measuring distances between objects and the vehicle based on the reflection of light. It is also helpful for more accurate real-time localization. Additionally, radar and ultrasonic sensors are also used to detect objects by electromagnetic pulses and ultrasonic pulse waves.

\subsection{The perception layer}
In the perception layer, semantic information is extracted from raw data by algorithms such as optical flow~\cite{agarwal2016review} and deep learning models. Currently, image data from cameras and cloud point data from LiDAR are widely used by deep learning models in the perception layer for various tasks such as localization, object detection and semantic segmentation.

\subsubsection{Localization}
Localization plays a critical role in the route planning task in an ADS. By leveraging localization technologies, the autonomous vehicle is capable of obtaining its accurate location on the map and understand the real-time ambient environment. Currently, localization is mostly implemented by the fused data from GPS, IMU, LiDAR point clouds, and HD map. 
Specifically, the fused data is used for odometry estimation and map reconstruction tasks. These tasks aim to estimate the movement of an autonomous vehicle,
reconstruct the map of the vehicle's surroundings, and finally determine the current location of the vehicle. In~\cite{DeepVO}, CNN and RNN were used to estimate the movement and poses of 
a vehicle, through continuous images taken by 
a camera. In~\cite{CodeSLAM}, a deep autoencoder was applied to encode observed images into a compact format for map reconstruction and localization.

\subsubsection{Road object detection and recognition}
Road object detection is a key issue for autonomous vehicles owing to the complexity of detecting large amounts of objects with different shape such as lanes, traffic signs, other vehicles, and pedestrians correctly in real time and ever-changing surrounding environments.
In the object detection field, Faster RCNN \cite{girshick2015fast} is considered effective to detect objects in images. 
You Only Look Once (YOLO) \cite{redmon2016you} is another famous object detection algorithm that converts the detection task to a regression issue. 
Currently, LiDAR-based object detection deep learning models are studied extensively by both researchers and industry practitioners. VoxelNet~\cite{zhou2018voxelnet} is the first end-to-end model that directly predicts objects based on LiDAR point cloud. PointRCNN~\cite{shi2019pointrcnn} adapts the architecture of RCNN to take 3D point cloud as input for object detection and achieves a superior performance.  

\subsubsection{Semantic segmentation}
Semantic segmentation in autonomous driving semantically segments different parts of an image into specific classes such as vehicles, pedestrians and ground. It is helpful for localizing the vehicle, detecting objects, marking lanes and reconstructing the map. In the semantic segmentation field, Fully Convolutional Network (FCN) \cite{long2015fully} is a basic deep learning model able to achieve good performance, which essentially modifies the fully connected layer in a normal CNN to convolutional layer. Also, PSPNet~\cite{zhao2017pyramid} is a famous semantic segmentation network that applies a Pyramid pooling architecture to better extract information from images.

\subsection{The cloud service}
The cloud server is commonly used as a service provider for many resource-reliant services in the autonomous driving field. First, a prior HD Map, which could be deployed at the cloud, is constructed by autonomous driving companies using LiDAR as well as other sensors. The HD Map contains much valuable information like road lanes, signs and obstacles. Therefore, the vehicle could use such data to initiate the pre-route planning and enhance the perception of the surrounding environment. Meanwhile, real-time raw data and perception data of other autonomous vehicles could be uploaded to the cloud by Vehicle to Everything (V2X) service to help keep HD Maps up-to-date, enabling HD Maps to provide more relevant real-time information such as surrounding vehicles on the same road. On the other hand, all deep learning models applied in an autonomous vehicle are first trained on the cloud in a simulation environment. When these models are verified, the cloud provides Over-the-Air (OTA) update to upgrade their software and deep learning models in autonomous vehicles remotely.     

\subsection{The decision layer}
\subsubsection{Path planning and object trajectory prediction}
Path planning is considered as a basic task for autonomous vehicles with respect to deciding a route between a start location and the desired destination and the object trajectory prediction task requires autonomous vehicles to predict trajectories of perceived obstacles with the help of sensors and perception layer. Recently, some researchers have tried to use Inverse Reinforcement Learning in order to achieve a superior results in path planning. By learning reward functions from human drivers, the vehicle is trained to be capable of generating a route more like a human being~\cite{gu2016human}. For trajectory prediction, some variations of RNN and LSTM~\cite{ gupta2018social} are proposed to achieve high prediction accuracy and efficiency. 
In addition, 3D spatial-temporal data and single CNN are tried by Luo~at~al. to forecast car trajectories~\cite{luo2018fast}. 

\subsubsection{Vehicle control via deep reinforcement learning}
Traditional rule-based algorithms cannot simply cover all complex driving scenarios. Deep reinforcement learning that trains an agent to learn how to act under different scenarios is thus more promising in autonomous driving scenarios. In~\cite{wulfmeier2016watch}, a CNN-based Inverse Reinforcement Learning model to plan a driving path using 2D and 3D data collected in many normal driving scenarios was proposed. In \cite{wolf2017learning}, a DQN based RL model was proposed for autonomous driving steering control. 

\subsubsection{End-to-End driving}
An E2E driving model is a special deep learning model that combines the perception and decision processes. In this scenario, the model predicts the current steering angle and driving speed based on the ambient sensing information. In~\cite{bojarski2016end}, a CNN architecture E2E driving model called DAVE-2 system takes front-face camera images as the input and predicts the current steering angle.

\section{Attacks in ADSs}
\label{Sec:threat model}
\begin{figure*}[tb!]
\centering

\includegraphics[width = \textwidth]{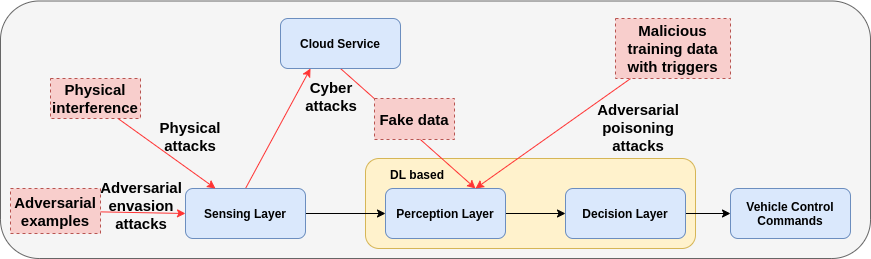}

\caption{An overview of attacks on each part in an ADS}
\label{fig:attack_overview}
\end{figure*}

In this section, we introduce various attacks towards ADSs in detail. Figure~\ref{fig:attack_overview} demonstrates the overview of attacks on each part in an ADS, which will be introduced in detail in this section. Table~\ref{tab: physical_attack} and~\ref{tab: threat} summarize both physical and adversarial attacks on ADSs.

\subsection{Physical attacks on sensors}
The sensing layer, commonly considered as the frontier layer of an ADS, is naturally seen as an attack target by adversaries. Attackers intend to degrade the quality of sensor data by adding noise signals or making sensors collect fake data by counterfeiting data signals. The low-quality or even fake data would affect the performance of deep learning models in the perception layer and the decision layer and further influence the behaviors of an autonomous vehicle. In this threat model, adversaries are assumed with a certain knowledge of hardware and specification of sensors applied on an autonomous vehicle, but they do not need to know details of deep learning models in other layers. Therefore, physical attacks on the sensing layer could be seen as black-box attacks on the deep learning-based ADS.  

For physical attacks on sensors, attackers could disturb the data collected from sensors or fabricate signals to fool sensors using some external hardware. There are two most common physical attacks in this context, namely jamming attacks and spoofing attacks.

\subsubsection{Jamming attack}
The jamming attack is deemed as the most basic physical attack that uses specific hardware to add noises into the environment to degrade sensors' data quality so as to make objects in the environment undetectable. 

Jamming attacks on a camera was experimented in~\cite{petit2015remote} and the camera is blinded by emitting the intense light into it. When the camera receives a much stronger incoming light than the normal environment, the auto-exposure function of the camera will not work normally, and the captured images would then be overexposed and not recognizable by deep learning models in the perception layer. In the experiment, front/side attacks with different distances and different light intensities were set. The results show that blinding attacks at a short distance in the dark environment setting could severely damage the quality of the captured images, which means that the perception system is not capable of recognizing objects effectively when such attack occurs. Another blinding attack was experimented in~\cite{yan2016can}, where attackers used a laser to cause temperature damages on cameras.
A blinding attack for LiDAR was proposed in~\cite{shin2017illusion}, in which the LiDAR is exposed under a strong light source that has the same wavelength as the LiDAR. Then, the LiDAR failed on perceiving objects from the direction of the light source. Jamming attacks on ultrasonic sensors and radars were experimented in~\cite{yan2016can}, where a roadside attack is launched through an ultrasound jammer to attack the parking assistance system of four vehicles. The results showed that under jamming attacks, the vehicles were incapable of detecting the surrounding obstacles. To attack the radar, a signal generator and frequency multiplier were used to generate electromagnetic waves against the Tesla Autopilot system in which the Autopilot system was also compromised. A jamming attack on ultrasonic sensors was simulated in~\cite{lim2018autonomous}. It was shown that other opposite placed ultrasonic sensors could substantially interfere the readings of the target ultrasonic sensor. In~\cite{son2015rocking}, the sound noise attacks on Gyroscopic sensors were launched, which were heavily used in the Unmanned Aerial Vehicles (UAV), leading to the fall of one UAV. In~\cite{kar2014detection}, GPS signals were found that they were vulnerable towards attacks from GPS jamming devices capable of producing large radio noises, which could adversely affect the navigation system.                 

\subsubsection{Spoofing attack}
Spoofing is a type of attack where adversaries use hardware to fabricate or inject signals during sensors data collection phase. The forged signal data could affect the perception of the environment and further cause abnormal behaviours of autonomous vehicles. In~\cite{petit2015remote}, a spoofing attack on LiDAR was tested. Specifically, as LiDAR could distinguish different objects at different positions by listening reflections of light achieving objects and echoing back, the counterfeit signal could return back ahead of the real signal. Consequently, LiDAR received with the counterfeit signal would lead to the wrong distance calculation between the vehicle and the object. Based on this idea, in the experiment, the real output signal of a wall was delayed and a counterfeited signal of the wall was created to produce the wrong distance information and successfully made LiDAR detect objects at the wrong distance. In~\cite{park2016ain}, a spoofing attack against LiDAR was implemented by injecting deceiving physical signals into the victim sensor, which makes the LiDAR ignore legitimate inputs. Similarly, ultrasound pulses and radar signals fabricated~\cite{yan2016can} to attack ultrasound sensors and a radar. GPS is another victim under the threat of spoofing attacks. In 2013, a yacht encountered a GPS spoofing attack, causing it to deviate away from the pre-set route~\cite{psiaki2016protecting}. In~\cite{MengHXLe19}, an open-source GPS spoofing generator was proposed, which can block all the legitimate signals. In~\cite{warner2002simple}, a similar GPS spoofing device was implemented to successfully attack commercial civilian GPS receivers. In~\cite{zeng2018all}, a GPS spoofing attack designed specifically for manipulating the navigation system was proposed. A GPS spoofing device, which could slightly shift the GPS location and then further manipulate the routing algorithm of the navigation system, was implemented. Subsequently, the autonomous vehicle would deviate from the original route. In addition to the attacks towards sensors, there are also spoofing attacks on cameras. In~\cite{davidson2016controlling}, a spoofing attack, aiming at the optical-flow sensing of UAV, was proposed. Attackers could alter the appearance of the ground plane, which would be captured by optical-flow cameras. Then altered images could adversely influence how the algorithms process the optical-flow information, and attackers could take over the control of the UAV by adopting this simple approach. There is another type of spoofing attack called relaying attack that usually occurs on LiDAR, aiming to deceive the target sensors by re-sending the original signal again from another position. The experiment in~\cite{petit2015remote} showed that two ghost walls in different locations were detected by LiDAR because of relaying attacks.   
In~\cite{MobilBye}, a projector was used to project spoofed traffic signs on cameras of 
a vehicle to make the vehicle interpret spoofed traffic signs as real signs.

\begin{table*}[]
\caption{Physical attacks on ADSs}
\label{tab: physical_attack}
\centering
\scalebox{1}{
\begin{tabular}{|c|l|l|l|l|}
\hline
\multicolumn{1}{|l|}{\textbf{Attack}}                                               & \textbf{Target sensor}                                                    & \textbf{Action}                                                                                          & \textbf{Implication}                                                                                                        & \textbf{Examples}                                                                                                                                                                                                      \\ \hline
\multirow{7}{*}{\textbf{\begin{tabular}[c]{@{}c@{}}Jamming\\ attack\end{tabular}}}  & Camera                                                                    & Extensive light blinding attack                                                                          & \begin{tabular}[c]{@{}l@{}}Make images overexposed and not recognizable;\\ Cause temperature damage on cameras\end{tabular} & \cite{petit2015remote}                                                                                                                                                                                \\ \cline{2-5} 
                                                                                    & LiDAR                                                                     & \begin{tabular}[c]{@{}l@{}}Blinding attacks by strong light\\ with same wavelength as LiDAR\end{tabular} & \begin{tabular}[c]{@{}l@{}}LiDAR cannot perceive objects from \\ the direction of light source\end{tabular}                 & \cite{shin2017illusion}                                                                                                                                                                             \\ \cline{2-5} 
                                                                                    & Ultrasonic sensor                                                         & Ultrasonic jamming device                                                                                & Obstacles cannot be detected                                                                                                & \cite{yan2016can}                                                                                                                                                                                     \\ \cline{2-5} 
                                                                                    & Ultrasonic sensor                                                         & \begin{tabular}[c]{@{}l@{}}Putting another ultrasound sensor\\ opposite to the target one\end{tabular}   & \begin{tabular}[c]{@{}l@{}}Both ultrasonic sensors cannot collect\\ accurate data\end{tabular}                              & \cite{lim2018autonomous}                                                                                                                                                                              \\ \cline{2-5} 
                                                                                    & Radar                                                                     & Generating electromagnetic waves                                                                         & Detected obstacles are disappeared                                                                                          & \cite{yan2016can}                                                                                                                                                                                     \\ \cline{2-5} 
                                                                                    & Gyroscopic  sensor                                                        & Sound noise                                                                                              & An UAV fall down                                                                                                            & \cite{son2015rocking}                                                                                                                                                                                 \\ \cline{2-5} 
                                                                                    & GPS                                                                       & GPS jamming device                                                                                       & Navigation system cannot work normally                                                                                      & \cite{kar2014detection}                                                                                                                                                                               \\ \hline
\multirow{5}{*}{\textbf{\begin{tabular}[c]{@{}c@{}}Spoofing\\ attack\end{tabular}}} & LiDAR                                                                     & \begin{tabular}[c]{@{}l@{}}Relaying signals of objects from\\ another position\end{tabular}              & LiDAR detects 'ghost' objects                                                                                               & \cite{petit2015remote}                                                                                                                                                                                \\ \cline{2-5} 
                                                                                    & \begin{tabular}[c]{@{}l@{}}LiDAR, Radar;\\ Ultrasonic sensor\end{tabular} & Fabricating fake signals                                                                                 & \begin{tabular}[c]{@{}l@{}}Sensors detect objects at wrong distance;\\ LiDAR ignores legitimate objects \end{tabular}                                                                              & \begin{tabular}[c]{@{}l@{}}\cite{petit2015remote}, \cite{yan2016can}
                                      \\ \cite{park2016ain}                                                \end{tabular}                                                                                                 \\ \cline{2-5} 
                                                                                    & GPS                                                                       & \begin{tabular}[c]{@{}l@{}}Using GPS-spoofing device\\ to inject fake signals\end{tabular}               & Navigation system is manipulated                                                                                            & \begin{tabular}[c]{@{}l@{}}\cite{psiaki2016protecting}, \cite{zeng2018all}, \\ \cite{MengHXLe19, warner2002simple}\end{tabular} \\ \cline{2-5} 
                                                                                    & \begin{tabular}[c]{@{}l@{}}Optical-flow \\ camera\end{tabular}            & \begin{tabular}[c]{@{}l@{}}Altering the appearance of \\ ground plane\end{tabular}                       & UAV is taken over                                                                                                           & \cite{davidson2016controlling} \\ \cline{2-5} 
                                                                                    
                                                                                      & \begin{tabular}[c]{@{}l@{}}Camera\end{tabular}            & \begin{tabular}[c]{@{}l@{}}Using a projector to project \\deceptive traffic signs onto ADAS\end{tabular}& \begin{tabular}[c]{@{}l@{}}The vehicle recognized the deceptive\\ traffic signs as real signs\end{tabular}                                                                                                          & \cite{MobilBye}
                                                                                      \\ \hline
\end{tabular}}
\end{table*}

\subsection {Cyberattacks on cloud services}
The cloud could be the target for many attacks from adversaries' perspective because of continuous communications between the cloud and autonomous vehicles, consequently resulting in instability of autonomous vehicles. 

Note that an HD Map could be updated in real time by information from other vehicles via V2X. This process is possibly controlled by attackers. For instance, Sybil attacks \cite{sinai2014exploiting} and message falsification attacks \cite{sinai2014exploiting} are designed to interfere the efficiency of the automatic navigation. Precisely, Sybil attacks focus on the real-time HD map updating in V2X, creating a large number of ``fake drivers'' in the target location system with fake GPS information. These attacks are designed to delude the system through the traffic jam and further interferes localization and navigation tasks in the vehicle. For the message falsification attacks, they intercept and tamper the traffic information updated from vehicle to the HD map server and spoofs other vehicles when updating the HD map information through this server.

Traditional cloud attacks are threatening the V2X network in which autonomous vehicles are connected to exchange information. Both Denial of Service (DoS) and Distributed DoS (DDoS)~\cite{long2005denial, du2020An} could cause the exhaustion of service resources, resulting in high latency or even the network unavailability of the V2X network. In this situation, autonomous vehicles may not be able to connect to the HD map for accurate navigation and perception service, which substantially endangers the safety of the autonomous vehicles.

One variation of attack aims at the over-the-air (OTA) channel in the cloud, where attackers could hijack the data transfer channel between the cloud and an autonomous vehicle and inject the malware into the vehicle~\cite{othmane2015survey}.

However, as attacks for cloud services are more relative to cyberattacks, we will not detail on such attacks and corresponding defending methods in this survey. 

\subsection {Adversarial attacks on deep learning models in perception and decision layers}
Recent research shows that deep learning models are particularly vulnerable to adversarial examples that add imperceptible noises on original input images. Even though adversarial examples look similar to normal images from human's view, they could mislead deep learning models to produce wrong predictions. By definition, an adversarial attack is a type of attack to construct such adversarial examples. Therefore, adversarial attacks pose considerable threats to ADSs due to the widespread usage of deep learning models in both the perception layer and the decision layer. 

In this section, we first introduce the definition of adversarial attacks along with some relevant concepts. Then we summarize the literature review the progress of adversarial attacks on different deep learning models in ADSs.

\subsubsection{Introduction to adversarial attacks}
Depending on attackers' ability, adversarial attacks could be categorized as either \textit{white-box} or \textit{black-box} attacks. In white-box attacks, attackers are assumed to know all the details of the target deep learning model including training data, neural network architecture, parameters, and hyper-parameters, while having the privilege to visit the gradients and results of the model at run time.

There are two types of adversarial attacks, i.e., adversarial evasion attacks occurring at the model inference time and poisoning attacks that happen in the model training period. Adversarial evasion attacks to deep learning models are first investigated for image classification tasks. Given a target deep learning model \(f\) and an original image \(x \in \mathcal{X}\) with its class \(c\), an adversarial attack could construct a human imperceptible perturbation \(\delta\) to form an \textit{adversarial example} \(x' = x + \delta \), which could delude the model to make a wrong prediction \(c' \neq c\). 

Commonly, there are three different kinds of white-box methods to generate adversarial examples, namely, gradient-based methods, optimization-based methods and generative model-based methods. 
\begin{itemize}
\item Gradient-based methods: These attack methods~\cite{goodfellow2014explaining, kurakin2016adversarial, kurakin2016adversarialphysical, tramer2017ensemble} are based on the Fast Gradient Sign Method (FGSM), as shown in Equation~(\ref{eq:FGSM}), to directly generate adversarial examples by adding the gradients of loss with respect to each pixel on original images~\cite{goodfellow2014explaining}.
            \begin{equation}
            \label{eq:FGSM}
                x' = x + \epsilon \, sign(\nabla J_\theta(x, c) )
            \end{equation}
\item Optimization-based methods: These attack methods~\cite{szegedy2013intriguing, carlini2017towards, moosavi2016deepfool} solve an optimization problem as
\begin{equation}
\label{opt_eq}
    \argmin_{x'} \alpha ||x-x'||_p + \ell(J_{\theta, \textbf{c}'}(x'))
\end{equation}
where the first part is the \(L_p\) distance between an original image and an adversarial image, and the second part is the constraint on the loss of the adversarial image~\cite{liu2016delving}. By solving this optimization problem, one could generate an adversarial image \(x'\) that is close to \(x\) in \(L_p\) distance but be classified as \(c'\).

\item Generative model-based methods: This type of attack~\cite{poursaeed2018generative, xiao2018generating} leverages generative models to generate targeted adversarial examples from original images. These methods normally learn a generative model \(\mathcal{G}\) by optimizing an objective function as
\begin{equation}
\label{generative_model}
    \mathcal{L} = \mathcal{L}_\mathcal{Y} + \alpha \mathcal{L}_{\mathcal{G}}
\end{equation}
 where \(\mathcal{L}_\mathcal{Y}\) denotes the cross-entropy loss between classification of adversarial examples and targeted class, and \(\mathcal{L}_{\mathcal{G}}\) measures the similarity between adversarial examples and original images.
\end{itemize}

 When it comes to the black-box attacks, attackers are assumed not having prior knowledge of the target model, but they can query the model and obtain the output of the model unlimitedly. There are also three different approaches to generate black-box adversarial examples:
\begin{itemize}
    \item Transfer-based methods: It was discovered that adversarial images targeted on a specific model were also found effective when dealing with other deep learning models, and this attribute is called the transferability of adversarial examples~\cite{liu2016delving}. Therefore, attackers could implement a similar model based on the input and output derived from the target model, and then initiate white-box attacks on their own model instead. The adversarial examples constructed based on their own model could be utilized to attack the target black-box model.
    
    \item Score-based methods: Although gradients information in a black-box model cannot be directly retrieved, the value of gradients could be estimated based on the probability score output of the target model and then used to craft adversarial examples~\cite{chen2017zoo}.
    
    \item Decision-based methods: These methods only rely on the final decision (e.g., top-1 classification result) of the target model to craft adversarial examples based on a randomly generated large perturbation and then iteratively reduce the perturbation while keeping adversarial features~\cite{Brendel2018Decision}.  
\end{itemize}

For attacks on ADSs, black-box attacks are more realistic. In addition, attacks on ADSs should occur in the physical world where sensors collect environmental information (e.g. images and point clouds) from different angles, light conditions, and distances. Therefore, this paper intends to cover physical black-box evasion attacks experimented in both simulation environment and the real world.

\subsubsection{Adversarial evasion attacks on ADSs}

This section first reviewed related attacks that were experimented in simulation environments, either real-world recording data or in simulated real-world scenarios. In addition, research experimented in the real world was also reviewed, which showed the harm of adversarial evasion attacks on ADSs in real life.  

\begin{figure}[tb!]
\centering
\includegraphics[width = .45\textwidth]{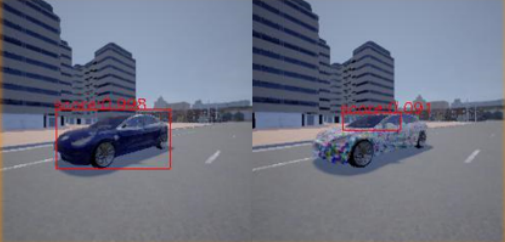}
\caption{Left: The vehicle can be detected normally; Right: The vehicle with adversarial texture cannot be recognized (image credit:~\cite{wu2020Physical})}
\label{Fig:texture attack}
\end{figure}

In~\cite{zhou2018deepbillboard}, an approach called \textit{DeepBillboard} to attack end-to-end driving models was proposed by replacing the original billboards on the roadside with adversarial perturbations. Specifically, the adversarial billboards were generated by the aforementioned optimization-based method. The method was tested on two end-to-end driving models on three datasets, along with different scenarios where billboards are positioned at different positions and angles. The result showed that their attack could make steering angle predictions deviate at most 23 degrees. In~\cite{Boloor2020attacking}, a Bayes Optimization-based approach was proposed to generate the painting of black lines on the road to counterfeit lane lines and make the vehicle deviate from the original orientation. Experiments were conducted in CARLA simulator~\cite{carla}, and results showed that E2E driving models were attacked and deviated to the orientation chosen by attackers. An updated approach proposed in~\cite{yang2020Finding} applied gradient-based optimization algorithm again to achieve quicker generation of black lines with higher deviation. In~\cite{wu2020Physical}, a decision-based approach was proposed to search and craft adversarial texture of vehicles. The average prediction score and the precision of object detectors in ADSs decreased sharply when presenting vehicles with adversarial texture (shown as Figure~\ref{Fig:texture attack}). Apart from that, some works also investigate attacks on LiDAR-based object detection in the simulation environment. In~\cite{cao2019adversarial}, a white-box optimization-based method was proposed to generate adversarial points and demonstrated how to inject these points into the original point cloud of an obstacle through laser. Experiments were conducted using LiDAR sensor data through a simulator released by Baidu Apollo. Experiment results showed that the average success rates of the attack were up to 90\%, while the number of injected adversarial points was larger than 60. The first black-box attack on LiDAR was proposed in~\cite{sun2020towards}, aiming to insert attack traces into point clouds to baffle LiDAR-based object detector. The experiment result on KITTI dataset achieved mean success rate at around 80\%.

In addition to research conducted under simulation environments, others study adversarial evasion attacks in the real world. For instance, an approach called ShapeShifter was proposed in~\cite{chen2018shapeshifter} to attack object detection model Faster R-CNN. The adversarial perturbation was generated by solve an optimization problem named Expectation over Transformation that aims to create a robust perturbation when it is captured from different angles with different lighting conditions. In the experiment, traffic signs with adversarial perturbations were printed in real world. The targeted attack success rate and the non-targeted one were reported to be $87\%$ and $93\%$, respectively. In~\cite{eykholt2018robust}, a method to generate robust physical perturbations was proposed. In the experiments, attackers could print the perturbed road signs and replace the true road signs with the perturbed ones ({\em subtle poster attacks}), or only print the physical perturbations as stickers with different colors and attach them on road signs ({\em camouflage abstract art attacks}). In the road test, success rates for the subtle poster attacks and camouflage abstract art attacks reached $100\%$ and $84.8\%$, respectively, both of which used a CNN model called LISA-CNN~\cite{mogelmose2012vision}.
\begin{figure}[tb!]
\centering
\includegraphics[width = .5\textwidth]{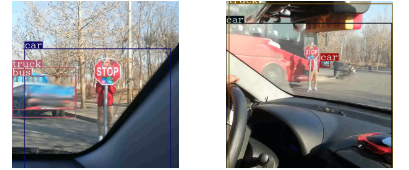}
\caption{The stop sign with an adversarial sticker cannot be recognized from different distance and angles (image credit:~\cite{zhao2019seeing})}
\label{Fig:stop sign attack}
\end{figure}
A generative model-based approach called Perceptual-Sensitive GAN was also proposed in~\cite{liu2019perceptual} in which an attention model was incorporated into the GAN to generate adversarial patches. The experiments conducted based on the physical world in a black-box setting showed that attacks could reduce classification accuracy from 86.7\% to 17.2\% on average. 
Similarly, methods proposed in~\cite{zhao2019seeing} can generate robust adversarial stickers to attack object detectors in two modes: Hiding attack that makes detectors fail to detect objects, and Appearing attack that makes detectors recognize wrong objects. Besides object detectors, E2E driving models were attacked in the physical world settings as revealed in~\cite{kong2020PhysGAN}. A method called PhysGAN was proposed to generate realistic billboard similar to the original one, but it could make autonomous vehicles deviating from their original route. The experiment results showed that billboards generated by PhysGAN could deviate steering angle predictions of E2E driving models up to 19.17 degrees.  

\begin{figure}[tb!]
\centering
\includegraphics[width = .45\textwidth]{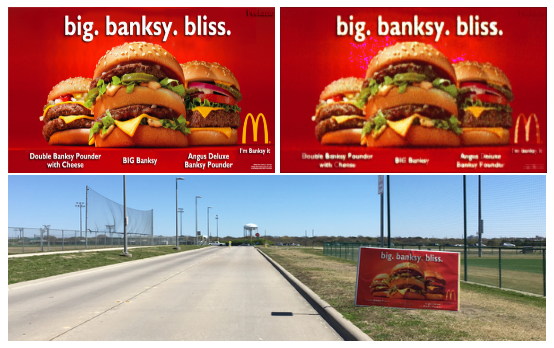}
\caption{Top Left: original billboard; Top Right: adversarial billboard generated by PhysGAN; Bottom: placing adversarial billboard in real world (image credit:~\cite{kong2020PhysGAN})}
\label{Fig:billboard attack}
\end{figure}

\subsubsection{Adversarial poisoning attacks on ADSs}
Poisoning attacks also fall into the types of adversarial attacks. More specifically, a poisoning attack works by injecting malicious data with triggers and misleading labels into original training data to make models learn specific patterns of triggers. During inference time, models are induced to produce wrong predictions when inputs contain malicious triggers. The poisoning attack is also considered to resemble the Trojan attack or backdoor attack.      
In \cite{liu2017trojaning}, the Trojan attack for E2E driving models was simulated. Adversarial triggers such as a square or an Apple logo were constructed and put on the corner of original input images. Experiment results showed that if the road images contained these malicious triggers, the vehicle could easily deviate from the pre-planned track. In~\cite{rehman2019backdoor}, poisoning attacks were conducted with four different triggers on four traffic sign recognition datasets, where one specific class of traffic signs was targeted. The experiment results showed that the CNN model could learn patterns of triggers and achieve more than 95\% accuracy on those poisoned images when the ratio of poisoned images was more than 5\%. Meanwhile, the overall accuracy of the total test dataset was more than 99\%, suggesting that it was difficult to determine if a model encounters poisoning attacks by only observing the results of test accuracy.
In~\cite{poison2019ding}, a poisoning attack on deep generative models like GAN for raindrop removing was proposed. Malicious data pairs were injected into original training data to make GAN learn the wrong map from the input domain to the output domain. The experiment result showed that when GAN removed raindrops, it simultaneously transformed red traffic light to green, or altered the number on speed limit sign.

\subsection{Analysis of attacks}

\noindent{\bf\em 1. Physical attacks are straightforward but limited in a certain range.}
Physical attacks on sensors could disrupt deep learning models by interfering the data collecting process. However, this type of attack requires the target in the proximity of the adversaries. For example, the camera blinding attack only occurs if the laser light is placed in front of the target vehicles, which makes such attacks difficult to implement. 

\noindent{\bf\em 2. Cyberattacks are harmful while challenging.}
Cyberattacks on the cloud could affect numerous autonomous vehicles connected in V2X network and thus result in severe consequences. However, for cyberattacks on the cloud, adversaries need to fabricate data transferred between the cloud and the vehicle or implement DDoS attacks by large Botnet. However, the encryption of data transmission process could hinder both attacks, and the cloud could deploy detection systems like~\cite{zhang2019multi, sun2020resilient} to defend DDoS attacks to some extent.

\noindent{\bf\em 3. Adversarial attacks are effective and pose threats in real world.}
Adversarial attacks, especially evasion attacks, would pose considerable risks to deep learning models in ADSs due to the existence of adversarial perturbations in the black-box setting. Table~\ref{tab: threat} shows some research works implemented black-box evasion attacks and experimented the effectiveness of their methods for attacking E2E driving models or object detectors in the perception layer of ADSs from different angles, distances, and light conditions in simulation environment or in the real world. For this line of attacks, adversaries could arbitrarily make malicious stickers and stealthily paste them everywhere. Adversarial poisoning attacks could occur in a scenario where corporate espionage has the chance to pollute training data, in which the attack could also be stealthy and hazardous. Therefore, it is essential to put a summary of current research on defenses against adversarial attacks. From the perspective of attacks, there may exist more powerful attacks to destroy autonomous vehicles, from which further research can be drawn.

\begin{table*}[]
\caption{Adversarial attacks on autonomous driving}
\label{tab: threat}
\centering
\scalebox{0.95}{
\begin{tabular}{|l|l|l|l|l|l|}
\hline
\multicolumn{1}{|c|}{Attack type}                                                        & \multicolumn{1}{c|}{Attack objective}     & \multicolumn{1}{c|}{Literature} & \multicolumn{1}{c|}{Method}                                                                                                                         & \multicolumn{1}{c|}{Attack setting} & \multicolumn{1}{c|}{Experiment setting}                                               \\ \hline
\multirow{11}{*}{\textbf{\begin{tabular}[c]{@{}l@{}}Envasion\\ \\ attacks\end{tabular}}} & \multirow{3}{*}{E2E driving model}        & \cite{zhou2018deepbillboard}                        & \begin{tabular}[c]{@{}l@{}}Replacing original billboard with adversarial billboard by\\ solving an optimization problem\end{tabular}                & White-box                           & Digital dataset                                                                       \\ \cline{3-6} 
                                                                                         &                                           & \cite{Boloor2020attacking}                        & \begin{tabular}[c]{@{}l@{}}Drawing black strips on the road by Bayesian\\  Optimization method\end{tabular}                                         & Black-box                           & \multirow{4}{*}{\begin{tabular}[c]{@{}l@{}}Simulation \\ \\ environment\end{tabular}} \\ \cline{3-5}
                                                                                         &                                           & \cite{yang2020Finding}                        & \begin{tabular}[c]{@{}l@{}}Drawing black strips on the road by Gradient-based\\  Optimization method\end{tabular}                                   & Black-box                           &                                                                                       \\ \cline{2-5}
                                                                                         & Object detection                          & \cite{wu2020Physical}                        & \begin{tabular}[c]{@{}l@{}}Drawing adversarial texture on other vehicles by\\  a discrete search method\end{tabular}                                & Black-box                           &                                                                                       \\ \cline{2-5}
                                                                                         & \multirow{2}{*}{3D Object detection}      & \cite{cao2019adversarial}                        & \begin{tabular}[c]{@{}l@{}}Generating adversarial points by optimization-\\ based method\end{tabular}                                               & White-box                           &                                                                                       \\ \cline{3-6} 
                                                                                         &                                           & \cite{sun2020towards}                       & Inserting attack trace into original point clouds                                                                                                   & Black-box                           & Digital dataset                                                                       \\ \cline{2-6} 
                                                                                         & \multirow{4}{*}{Traffic sign recognition} & \cite{chen2018shapeshifter}                        & \begin{tabular}[c]{@{}l@{}}Replacing true traffic signs with adversarial traffic signs \\ generated by solving an optimization problem\end{tabular} & White-box                           & Real world                                                                            \\ \cline{3-6} 
                                                                                         &                                           & \cite{eykholt2018robust}                       & \begin{tabular}[c]{@{}l@{}}Pasting adversarial stickers that generated by optimization-based \\ approach on traffic signs\end{tabular}              & White-box                           & Real world                                                                            \\ \cline{3-6} 
                                                                                         &                                           & \cite{liu2019perceptual}                        & Generating transferable adversarial patches by GAN                                                                                                  & Black-box                           & Real world                                                                            \\ \cline{3-6} 
                                                                                         &                                           & \cite{zhao2019seeing}                       & \begin{tabular}[c]{@{}l@{}}Generate transferable adversarial traffic signs and stickers by\\ Feature-interference reinforcement\end{tabular}        & Black-box                           & Real world                                                                            \\ \cline{2-6} 
                                                                                         & E2E driving model                         & \cite{kong2020PhysGAN}                      & Generate adversarial billboard by GAN                                                                                                               & White-box                           & Real-world                                                                            \\ \hline
\multirow{3}{*}{\textbf{\begin{tabular}[c]{@{}l@{}}Poisoning\\ \\ attacks\end{tabular}}} & E2E driving model                         & {[}64{]}                        & \multirow{2}{*}{Adding poisoning images with triggers into training data}                                                                           & White-box                           & \begin{tabular}[c]{@{}l@{}}Simulation \\ environment\end{tabular}                     \\ \cline{2-3} \cline{5-6} 
                                                                                         & Traffic sign recognition                  & \cite{rehman2019backdoor}                      &                                                                                                                                                     & White-box                           & Digital dataset                                                                       \\ \cline{2-6} 
                                                                                         & Rain drop removing                        & \cite{poison2019ding}                       & Adding poisoning image pairs with triggers into training data                                                                                       & White-box                           & Digital dataset                                                                       \\ \hline
\end{tabular}}
\end{table*}

\section{Defense methods}
\label{defense}
In this section, we take a close look at some existing defense methods against both physical attacks and adversarial attacks. We also briefly discuss about defenses for cloud services. The limitations of current defenses against adversarial evasion and poisoning attacks are further analyzed and summarized in Table~\ref{tab:defense}.

\subsection{Defense against physical sensor attacks}
Among all the countermeasures for physical sensor attacks, redundancy~\cite{petit2015remote, yan2016can, lim2018autonomous} is the most promising strategy to defend jamming attacks. Redundancy means that a number of the same sensors are deployed to collect a designated type of data and fuse them as the final input for the perception layer. For example, when attackers commit the blinding attack on one camera, others could still collect normal images for the environment perception. Undoubtedly, this method leads to more financial costs. Also, sensor data fusion is generally considered as an intractable research issue. To improve the robustness of cameras, another approach is that a near-infrared-cut filter is applied to filter the near-infrared light in the daytime to improve the quality of collected images~\cite{petit2015remote}, which is however unable to work effectively at night time. Alternatively, using photochromic lenses to filter a specific type of light is also an option to upgrade cameras. Subsequently, jamming attacks on these cameras could be mitigated. For ultrasonic sensors and radars, as noises hardly occur in a normal working environment, it is not difficult to build a detection system to detect the incoming jamming attacks~\cite{yan2016can}. Moreover, a jamming detection system for GPS was proposed in~\cite{kar2014detection}, which expedites GPS information from multiple sources such as roadside monitoring stations and mobile phones to improve the accuracy of GPS information. 

One effective way to defend spoofing attacks is to introduce randomness into data collection~\cite{petit2015remote, yan2016can}. For example, attackers could commit accurate attacks on LiDAR because there is a fixed probe window for LiDAR to receive signals. If the probe time is set to be random, it then becomes harder for adversaries to send fake signals. PyCRA is a spoofing detection system based on this idea~\cite{shoukry2015pycra}. 
Furthermore, data fusion mechanism is considered effective to defend spoofing attacks. Therefore, fusing data from cameras, LiDAR, radars and ultrasonic sensors could help stabilize the performance of the perception layer.  

There are some obvious limitations to the existing sensor attacks. For instance, many attacks require external hardware to generate noises and fake signals within a short distance near the target vehicle. A human may recognize the occurrence of attacks such as the camera blinding attack from the front of the vehicle and take over the vehicle to avoid accidents. Therefore, even if the development of autonomous vehicles achieves a highly automated level, it is still necessary to set a security guard in the vehicle as a guarantee.

\subsection{Defense for cloud services} 
In the V2X map updating process, the HD map needs to be secured for authenticity and integrity. Each map package should contain the unique identity of the service provider. The integrity and confidentiality should also be ensured during the updating to prevent stealing or changing of data. In~\cite{zeng2018all}, encryption and authentication are applied for GPS data during transmission to defend message falsification attacks. In \cite{mahmud2005secure}, a symmetric key encryption-based update technique was proposed to apply a link key between the service supplier and vehicles to form a secure package updating connection. In~\cite{nilsson2008secure}, a hash function-based update technique was proposed. This technique first divided the package into several data fragments and then created a hash chain of these data fragments in the decreasing order. Before the package being collected by the vehicle, elements in the hash chain were encoded using the pre-shared encryption key.

\subsection{Defense against adversarial evasion attacks}

Currently, many defenses against adversarial evasion attacks are proposed. In this survey, we reviewed existing defenses and divided them into different categories.
Adversarial defenses can be categorized into proactive and reactive methods. The former focuses on improving the robustness of the targeted deep learning models, while the latter aims to detect and counter adversarial examples before they are fed into models. There are five main types of proactive defense methods, namely, adversarial training, network distillation, network regularization, model ensemble, and certified defense. The primary reactive defenses are called adversarial detection and adversarial transformation. Though most of defenses have only experimented on image classification tasks, ideas of these defenses is a good generalization to other tasks in autonomous driving, considering the similar approaches in improving the robustness of models or pre-processing model inputs that are not limited on image classification. To validate whether these defenses can be applied in ADSs, we analyzed and compared them in Section~\ref{sec:discussion of defense}.

\begin{table*}[h!]
\centering
\caption{Summary of adversarial defenses} 
\label{tab:defense}
\scalebox{0.85}{
\begin{tabular}{|l|l|l|l|l|}
\hline
                                    & Name & Function & Example           & Analysis                                                                                                                           \\ \hline
\multirow{5}{*}{\textbf{\begin{tabular}[c]{@{}l@{}} \\ \\ Proactive\\ \\ \\ \\ defenses\end{tabular}}} & Adversarial training       & \begin{tabular}[c]{@{}l@{}}Train a new robust model based on new \\ dataset that involves adversarial examples.\end{tabular}               &  \cite{goodfellow2014explaining} \cite{kurakin2016adversarial} \cite{tramer2017ensemble}  & \multirow{2}{*}{\begin{tabular}[c]{@{}l@{}}Increasing time and resource consumption\\ for autonomous driving model training;\\ only effective for simple attacks\end{tabular}} \\ \cline{2-4}
                                       & Defensive distillation     & \begin{tabular}[c]{@{}l@{}}Train a new robust model by distilling hidden\\ layer information from the original model\end{tabular}          &      \cite{papernot2016distillation}             &   \\ \cline{2-5} 
                                       & Model ensemble & \begin{tabular}[c]{@{}l@{}}Ensemble multiple models for making the final \\prediction to improve the robustness \end{tabular} & \cite{kurakin2018adversarial} \cite{liu2018towards} \cite{pang2019Improving} &  Increasing resource consumption                                                           \\ \cline{2-5} 
                                      & Network regularization     & \begin{tabular}[c]{@{}l@{}}Train a robust model based on a new objective\\ function containing perturbation-based regularizer\end{tabular} &      \cite{yan2018deep}  \cite{gu2014towards} \cite{cisse2017parseval}  &  \multirow{2}{*}{\begin{tabular}[c]{@{}l@{}}Increasing time and resource consumption\\ for autonomous driving model training;\\ only effective for simple attacks\end{tabular}} \\ \cline{2-3}
                                       & Certified robustness     & \begin{tabular}[c]{@{}l@{}}Change the architecture of the model to make\\ it provably robustness against certain adversarial examples\end{tabular} &      \cite{lecuyer2019certified} \cite{raghunathan2018certified} \cite{wong2018Provable}         &
                                \\ \hline
\multirow{3}{*}{\textbf{\begin{tabular}[c]{@{}l@{}}Reactive \\ \\ defenses\end{tabular}}} & Adversarial detection      & \begin{tabular}[c]{@{}l@{}}Detect adversarial examples by a detector or \\ verifying the feature representation of inputs; \\Detect hijacked image with triggers or identify \\ poisoning attack in the model\end{tabular}     &           \begin{tabular}[c]{@{}l@{}} \cite{zheng2018robust}  \cite{lee2018simple}  \cite{xu2017feature} \\ \cite{gao2019strip} \cite{chen2018detecting} \cite{wang2019neural}  \end{tabular}           & \begin{tabular}[c]{@{}l@{}}Detector is not available if it requires much\\ resource\end{tabular}                                  \\ \cline{2-5} 
                                       & Adversarial transformation & \begin{tabular}[c]{@{}l@{}}Apply transformation to convert adversarial examples\\ back to clean images\end{tabular}                      &   \cite{guo2017countering} 
                                 \cite{samangouei2018defense}   \cite{shen2017ape} \cite{liao2018defense}  
                                 & \begin{tabular}[c]{@{}l@{}}They may reduce the performance of\\ autonomous driving models under normal conditions\end{tabular}
                                               \\ \hline
\end{tabular}}
\end{table*}
\subsubsection{Proactive defenses}
\textbf{Adversarial training} was initially proposed in \cite{goodfellow2014explaining}. This defense method targeted to re-train a more robust model on the dataset that combines original data and adversarial examples. In \cite{kurakin2016adversarial}, the experiment result showed that adversarial training was just useful to defend against one-step attacks that generate adversarial examples by only one-time operation. In \cite{tramer2017ensemble}, a method to combine multiple attacks together was proposed to generate adversarial examples for adversarial training. However, it failed to improve the robustness of models against unseen attacks.

\textbf{Defensive distillation} was proposed in~\cite{papernot2016distillation}. This defense method trained a new model by using probability logits information of the original model as the soft labels. The new model trained in this way is less sensitive to the change of gradients, so it is more robust against adversarial examples. However, a new optimization-based attacks was proposed in~\cite{carlini2017towards} to bypass the defense.

\textbf{Network regularization methods} train models against adversarial examples by adding another adversarial-perturbation based regularizer into the original objective function \cite{yan2018deep}.
 In \cite{gu2014towards}, contractive autoencoders were proposed and generalized into neural networks by using \(L_2\) norm of the layer-wise Jacobian matrices as the regularizer. In \cite{cisse2017parseval}, a parameter \(\alpha\) was introduced to control the overall Lipschitz constant of the whole model. Experiments on CIFAR-10/CIFAR-100 \cite{krizhevsky2009learning} showed that such network regularized models have higher robustness against FGSM attack than the original models.  
 
 \textbf{Model ensemble methods} were designed to improve the robustness by constructing an ensemble model that aggregates several individual models~\cite{kurakin2018adversarial}. In~\cite{liu2018towards}, a random self-ensemble approach was proposed to derive the final prediction results by averaging predictions over random noises that are injected into the model. This approach is equivalent to ensembling infinite number of noisy models. In~\cite{pang2019Improving}, an adaptive approach was proposed to train individual models with larger diversity. Then the ensemble of individual models could achieve better robustness because the attack is more difficult to transfer among individual models.

\textbf{Certified robustness methods} aim to provide provable defense against adversarial attacks with adversarial examples generated by several threat models~\cite{lecuyer2019certified, raghunathan2018certified, wong2018Provable}. In~\cite{lecuyer2019certified}, a method called \textit{PixelDP} was proposed as certified defense. This method adds an additional noise layer into the original model to serve the purpose of random perturbation whose size is smaller than a threshold on original inputs or features representations. The new model is then more robust against adversarial examples if the injected perturbations smaller than the pre-defined threshold.

\subsubsection{Reactive defense}
\textbf{Adversarial detection} could detect adversarial examples by introducing another classifier that could differentiate the feature representation of adversarial examples from natural images. In \cite{zheng2018robust}, an intrinsic-defender (I-defender) was proposed to identify adversarial examples from original images under unknown attack methods. I-defender explores an intrinsic property of the target model, e.g., the distribution of hidden states of normal training sets, and then uses the intrinsic property to detect adversarial examples. Similarly, an effective method for DNNs with the softmax layer was proposed in~\cite{lee2018simple} to detect abnormal samples including out-of-distribution (OOD) and adversarial examples. The idea was to use Gaussian discriminant analysis \cite{hastie1996discriminant} to measure the probability density of test samples on feature spaces of DNNs. In~\cite{xu2017feature}, an approach called Feature Squeezing was proposed to detect adversarial examples by squeezing the
color bit depth of each pixel. If the difference between the predictions on the original input and the squeezed input is over a threshold, the original input is more likely an adversarial example. 

\textbf{Adversarial transformation} is a set of approaches that could apply transformations on adversarial examples to reconstruct them back to clean images. In \cite{guo2017countering}, the effects of five image transformations for defending against FGSM, I-FGSM, DeepFool and C\&W attacks were investigated. The results showed that transformations were partially effective to defend against adversarial perturbations, while randomized (e.g., image cropping) and non-differentiable (e.g., total variation minimization) transformations were stronger defenses.
In \cite{samangouei2018defense}, a framework called defense-GAN was proposed, which learns the underlying distribution of the image dataset and can generate images falling in this distribution. When an adversarial example was fed into the target model, the defense-GAN generated many images that are similar to the adversarial example in \(L_2\) distance and then search the optimal one as the input of the target model. In~\cite{shen2017ape}, another GAN model called Adversarial Perturbation Elimination GAN (APE-GAN) was proposed to denoise adversarial examples, which uses adversarial examples \(X'\) as the input to directly output their corresponding denoised images \(\mathbf{G}(X')\). The experiments confirmed that APE-GAN is able to defend against common attacks.
In \cite{liao2018defense}, the High-Level Representation Guided Denoiser (HGD) was proposed to transform adversarial examples through an auto-encoder network.
The key idea of HGD is that it does not minimize the \(L_2\) distance between the generated image and original image but shortens the distance of feature representations at \(l\)-th layer of the target model \(f\).
The experiment result shows that HGD ranks first in NIPS adversarial defenses competition \cite{kurakin2018adversarial}.

\subsection{Defense against adversarial poisoning attacks}
A number of defense methods for poisoning attacks have been proposed in some recent research works. The general philosophy is to only detect whether the current input image is a hijacked image with triggers. Another high level thought is to identify the poisoning attack in the model and then remove the backdoor or Trojan. Both of the ideas belong to reactive adversarial detection defenses. In~\cite{gao2019strip}, a detection method called STRIP was proposed, which compared the predictions of the original input image and a perturbed input image that is generated by superimposing another clean image from training data. If the input image did not contain a trigger, the predictions of input image and perturbed image should be different. However, if the input image was deemed to contain a trigger, the predictions should be same because the perturbed image also contains the trigger that dominates the prediction of the model. In this manner, the hijacked image with a trigger could thus be detected. 

In~\cite{chen2018detecting}, a detection method was proposed to distinguish the clean input image from the malicious ones with a trigger. The method was based on an observation that even though clean images and hijacked images were classified to have the same label, the final output of the last activation layer is drastically different. Due to the observation, the method adopted a clustering algorithm to group the poisonous data owing to this difference. In~\cite{wang2019neural}, a comprehensive method was proposed to identify and mitigate poisoning attacks at the model level. Firstly, different triggers were created to attack each label, and the weights of neurons activated by the detected trigger were then removed to makes the trigger ineffective. The experiment results illustrated that this approach could significantly reduce attack success rates, even going down from over 90\% to 0\% for some poisoning attacks.

\subsection{Analysis of defenses}
\label{sec:discussion of defense}

{\bf\em 1). Defense against physical sensor attacks is costly but effective.} Redundancy defence requires to use numerous sensors in the same type to collect the target data and combines the data together before sending it to the perception layer. Even though it leads to a significant expenditure on the sensors, redundancy is considered as a simple yet effective way to defence jamming attack. Apart from the cost, the technical issue of data fusion also needs to be taken into account.

{\bf\em 2). Current adversarial defense methods are not suitable in autonomous vehicles.} Table \ref{tab:defense} summarizes the reviewed defense techniques. For proactive methods, adversarial training and defensive distillation need to train a new robust model following the original model training. However, the training of autonomous driving models generally requires large datasets and incurs a significant training time. Importing these techniques will undoubtedly result in the resource overhead. Moreover, adversarial training and defensive distillation are only effective when dealing with simple adversarial attacks like FGSM. As stated in the preceding section, model ensemble methods take advantage of results from multiple models to improve the robustness, which also cause large extra resource overhead. On the other hand, network regularization and robustness methods can be integrated into the training process of autonomous driving models without incurring large extra resource overhead. Yet, it is worth mentioning that such methods mostly experimented on DL models with simple network architecture, and its effectiveness needs to be further verified in ADS settings. For reactive methods, adversarial transformation process could achieve a satisfactory result when applying on adversarial examples. Still, the performance may degrade on normal inputs, which is unacceptable for safety-critical autonomous vehicles. When it comes to the adversarial detection, some techniques suggest to take advantage of other classifiers to detect adversarial examples, which is also infeasible as the classifier requires additional computation resources and might violate stringent timing constraints in ADSs.
Therefore, other adversarial detection methods that do not cause considerable resource overhead could be incorporated in autonomous driving models. Also, other techniques that are helpful for improving the robustness of autonomous driving models should be explored in the future. In addition, as autonomous driving is a real-time interactive process, thus real-time monitoring and defense are of great importance in order to keep the safety of autonomous vehicles.

\begin{figure*}[h!]
\centering

\includegraphics[width = .85\textwidth]{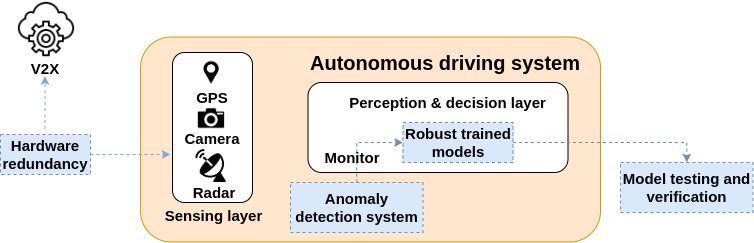}

\caption{Overview of defense framework on ADS}
\label{fig:defense_overview}
\end{figure*}
\section{Future Directions}
\label{Sec:future direction}
In this survey, we conduct a comprehensive review on some existing attacks, including both physical and adversarial ones, as well as corresponding defense methods along with the detailed analysis of their availability and limitation in the deep learning-based ADSs. 
This survey discusses various adversarial attacks that could be detrimental to deep learning autonomous driving models and identify relevant safety threats. In this section, we uncover further research directions for possible attacks on ADSs and strategies for improving the robustness of ADSs against adversarial attacks. In particular, we propose the potential detection mechanisms explicitly applicable for current autonomous vehicles to defend against adversarial attacks as the majority of existing adversarial defense methods are not designated for deep learning-based ADSs in the first place. 

\subsection{Potential attacks in future research}
\subsubsection{Adversarial attacks on the whole ADS}
\label{sec:potential_attack_1}
Most of existing attack-related research normally focus on single target (e.g., physical attacks on cameras or GPS) or a sub-task in an ADS (e.g., adversarial attacks on object detectors). Some research simplify an ADS as an E2E driving model for attacking. However, as an ADS is composed of several layers, and inputs from different sensors are tend to be fused at first to provide environment information. The success launch of attaching on one sensor or one deep learning model does not necessarily mean that it would effectively make the ADS produce wrong control decisions. For example, object detection in autonomous vehicle could be realized through the fusion of camera-based and LiDAR-based deep learning models, and only attacking either one of them may not affect the final recognition results. Therefore, it is essential to investigate attacks against models based on multi-modal inputs and attacks against full-stack ADS like Apollo and Autoware.

\subsubsection{Semantic adversarial attacks}
Currently, some research starts to investigate semantic adversarial attacks that focus on changing specific attributes such as light conditions and clarity of the input to generate natural adversarial examples.  
The existence of semantic adversarial attacks demonstrates that deep learning models tend to make mistakes in real-world even without adversary, meaning that weather, light, or other conditions could easily be turned into semantic adversarial attributes in coincidence. Such uncertainty would pose unexpected threats to autonomous vehicles. Therefore, the research of semantic adversarial attacks is necessary in terms of achieving better performance and robustness of deep learning models applied in ADSs.

\subsubsection{Reverse-engineering attacks}
Other than adversarial attacks, reverse-engineering attacks on ADSs are another possible research direction. For instance, an approach to construct a metamodel was proposed in~\cite{oh2019towards} for predicting attributes of black-box classifiers. Based on extracted attributes, adversarial examples could be created to attack black-box classifiers. Also, the parameters of a neural network could be recovered by using the side-channel analysis technique~\cite{batina2019csi}. Since deep learning models are now widely adopted in the industry, the valuable information contained in the structure and parameters should be protected securely. Simply put, the model should be robust enough against various reverse-engineering attacks to preserve both integrity and stability of the model.

\subsection{Strategies for robustness improvement}

Based on reviewed attacks and defenses, we propose a defense framework to improve the robustness of ADSs as shown in Figure~\ref{fig:defense_overview}. The framework could be applied as practice in industry. Specifically, we propose four strategies \textit{hardware redundancy}, \textit{robust model training}, \textit{model testing and verification}, and \textit{anomaly detection} that could be investigated in the future. 

\subsubsection{Hardware redundancy}
As discussed in Section~\ref{sec:potential_attack_1}, current attacks only focus on one specific target in ADSs, applying multiple sensors to perceive the environment hence is a good way to improve the robustness. In addition, with the development of V2X, an autonomous vehicle can receive information from roadside units like surveillance cameras or from other nearby vehicles. By fusing sensor data from V2X clients and data collected by sensors on the vehicle, the perceived environment information would be more robust against being turned into adversarial input.

\subsubsection{Model robustness training}
From the perspective of adversarial defense, training autonomous driving models that are naturally robust against adversarial examples is a promising research direction. For instance, network regularization follows this line of thought. However, many network regularization methods merely focus on specific adversarial examples. Recently, in \cite{zhang2019theoretically}, a new regularization method was proposed by introducing surrogate loss to improve the robustness of models. This method won the first place in the NeurIPS 2018 Adversarial Vision Challenge to defend adversarial examples. Another assuring approach to improve the robustness is to modify the network architecture of models. 

\subsubsection{Model testing and verification}
After the model training stage, it is also essential to apply viable testing and verification techniques on the trained models to measure their performance against adversarial examples. Data-driven deep learning models are vastly different from traditional software and thus difficult to benefiting from the existing software engineering test methods~\cite{murphy2007approach}. Currently, some testing and verification tools have been developed to cope with this issue. For example, in \cite{huang2017safety}, a white-box framework was proposed to exhaustively search adversarial examples. Therefore, applying testing and verification techniques to prevent the adversarial examples is another promising research direction.

\subsubsection{Adversarial attacks detection in real time}
Lastly, before deploying a robust ADS, sound adversarial attack detection and monitoring system are urgently needed as the last-line defense against a variety of attacks for autonomous vehicles in real time. Current adversarial attack detection methods usually rely on an auxiliary model to detect adversarial examples, which may not be feasible for the resource-constrained autonomous vehicles. Therefore, detecting abnormal behavior caused by adversarial examples without incurring the resource overhead is an important research direction. Adversarial detection techniques such as the one in ~\cite{lee2018simple} explored in Section~\ref{defense} do not introduce new models or layers into original autonomous driving models, they hence do not cause large overhead. However, these works were only experimented on the public datasets like MNIST and CIFAR-10. It is essential to conduct comprehensive experiments on datasets of real-world autonomous driving tasks. Another possible research direction is to deploy an anomaly detection system on the Cloud/Edge server to monitor and analyze the data uploaded by autonomous vehicles. The Cloud/Edge server has powerful computation so we could implement more accurate detection methods to detect adversarial examples. However, how to ensure the timely response, handle time synchronization and deal with a large amount of sensor data in an anomaly detection system at the running time remain unsolved.
In~\cite{li2019desvig}, a decentralized swift vigilance framework was proposed to recognize abnormal inputs with ultra-low latency.
In~\cite{zheng2016efficient}, a highly scalable anomaly detection mechanism was created to enable the gathering and compression of event data in a highly distributed environment, in which a desired balance between response time and accuracy is well achieved.

\section{Conclusion}
\label{Sec: conclusion}
The deep learning-based ADS is the key to realize a more intelligent self-driving system. However, the system is vulnerable to diverse attacks. In this survey, potential safe-threatening attacks are analyzed in the workflow of the deep learning-based ADS, including physical attacks, cyberattacks and adversarial attacks. The physical attack is straightforward but shows certain limits that could be dealt with defence methods effectively. The cyberattack is considered difficult to launch in large scale, while system defence methods are easy to implement. The adversarial attack is effective, and more defence methods against it are needed, as traditional defence methods are not suitable in the self-driving context. In future research, adversarial attacks on LiDAR and deep reinforcement models and reverse-engineering attacks are potential attacks must be researched. To improve the robustness of the ADS, model robustness training, model testing and verification and adversarial attacks detection in real-time should also be studied thoroughly.

\ifCLASSOPTIONcaptionsoff
  \newpage
\fi

\begin{IEEEbiography}[{\includegraphics[width=1in,height=1.25in,clip,keepaspectratio]{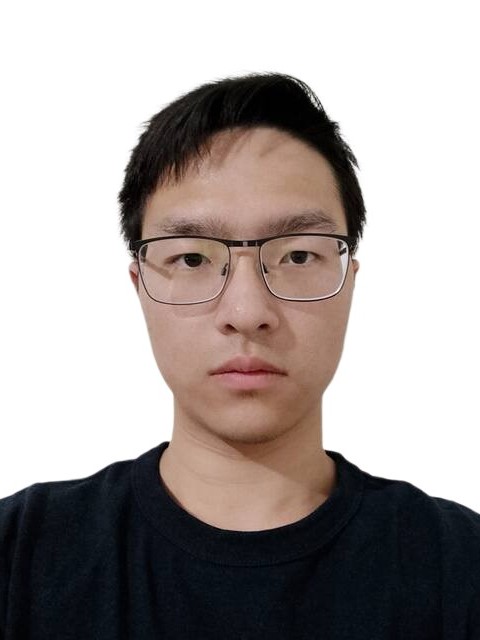}}]{Yao Deng}
(S'21) received the Bachelor degree of Information Technology from Deakin University, Australia in March 2018, the Bachelor degree of Software Engineering from Southwest University, China in July 2018, and the Master of Research degree in Computing from Macquarie University, Australia in 2020. He is currently pursuing his PhD degree at Macquarie University. His current research interests include adversarial attacks and defenses, metamorphic testing, simulation testing, and anomaly detection on autonomous driving systems.
\end{IEEEbiography}

\begin{IEEEbiography}[{\includegraphics[width=1in,height=1.25in,clip,keepaspectratio]{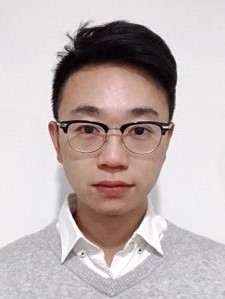}}]{Tiehua Zhang}
  
(S'21) received the B.S. degree from the School of Computer Science and Technology at Jilin University, China in 2013, the M.E. from the School of Computing and Information Systems at the University of Melbourne, Australia in 2015, and the Ph.D. degree from the School of Software and Electrical Engineering at Swinburne University of Technology, Australia in 2020, respectively. He then worked as a Postdoctoral Researcher in the Department of Computing at Macquarie University. From 2015 to 2017, he was a Software Engineer in Australia, focusing on industrial projects and solutions. He is currently an AI Specialist at Ant Group, China. His research interests include collaborative learning/optimization, the Internet of Things, fog computing, and edge intelligence.
\end{IEEEbiography}

\begin{IEEEbiography}[{\includegraphics[width=1in,height=1.25in,clip,keepaspectratio]{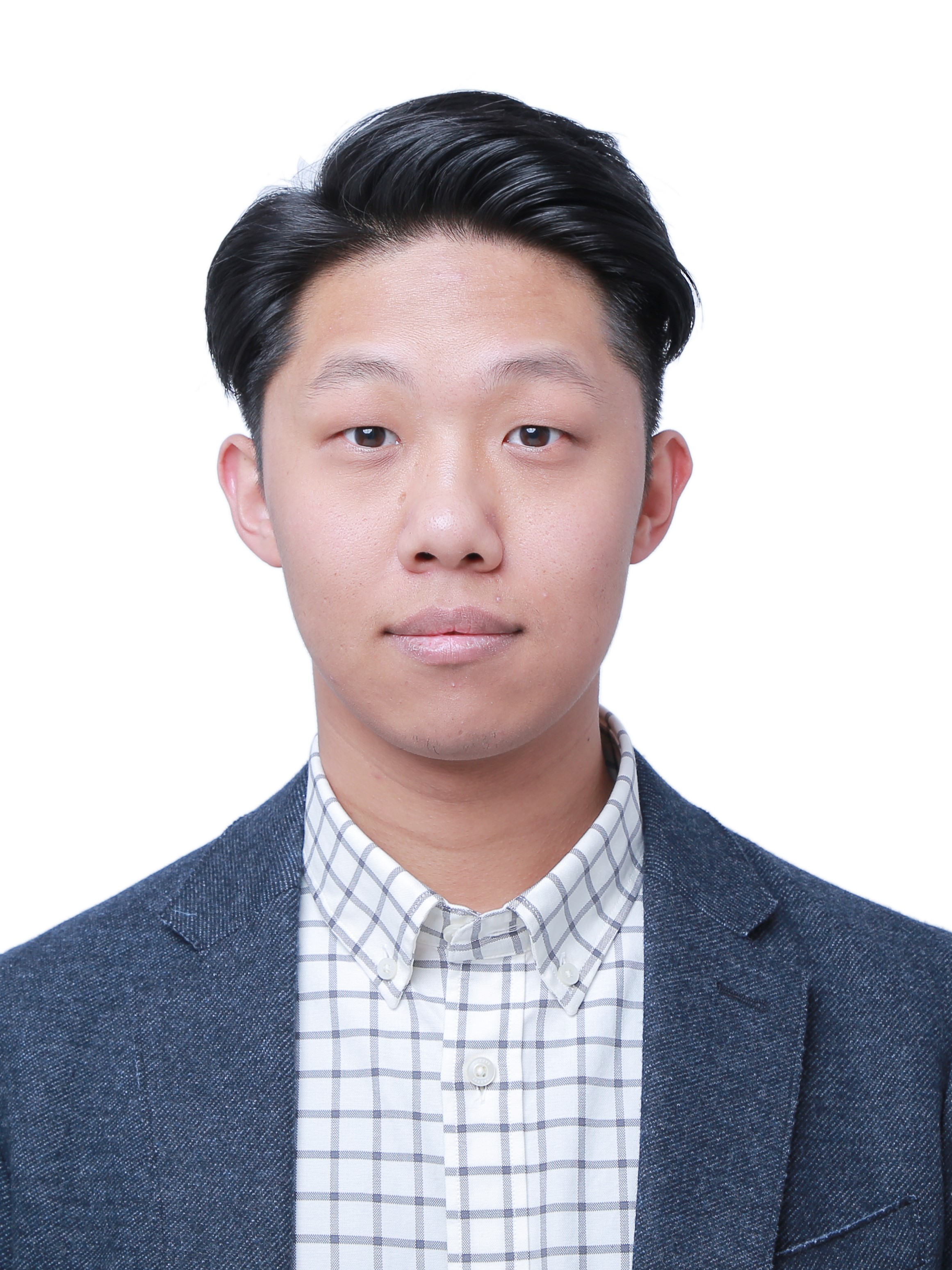}}]{Guannan Lou}
  
(S'21) received the Bachelor degree of Information Technology from Deakin University, Australia in March 2018, the Bachelor degree of Software Engineering from Southwest University, China in July 2018, and the Master of Data Science from Sydney University, Australia in 2020. Currently he is pursuing the PhD degree at Macquaire University. His research interests include machine learning, machine learning security, metamorphic testing and natural language processing.
\end{IEEEbiography}

\begin{IEEEbiography}[{\includegraphics[width=1in,height=1.25in,clip,keepaspectratio]{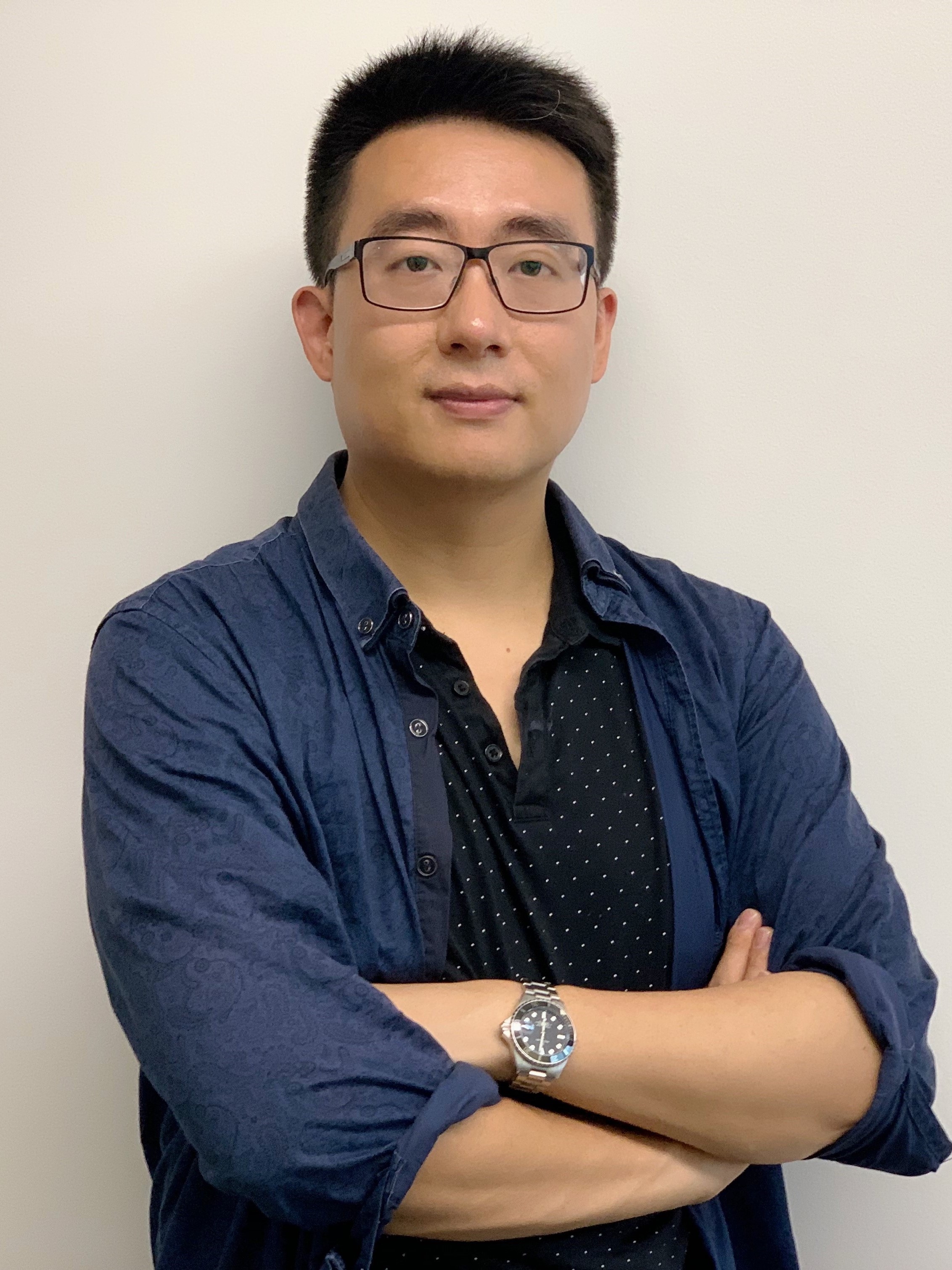}}]{Xi Zheng}
  
(M'12) received the Ph.D. in Software Engineering from UT Austin in 2015. From 2005 to 2012, he was the Chief Solution Architect for Menulog Australia. He is currently the Director of Intelligent Systems Research Group (ITSEG.ORG), Senior Lecturer (aka Associate Professor US) and Deputy Program Leader in Software Engineering, Macquarie University, Sydney, Australia. His research interests include CPS verification, machine learning security, human vehicle interaction, edge intelligence and intelligent software engineering. He has a number of highly cited papers and served as the PC member for IEEE International Conference on Pervasive Computing and Communications (PerCom) (CORE A*) and International Conference on Trust, Security and Privacy in Computing and Communications (TrustCom) (CORE A).
\end{IEEEbiography}

\begin{IEEEbiography}[{\includegraphics[width=1in,height=1.25in,clip,keepaspectratio]{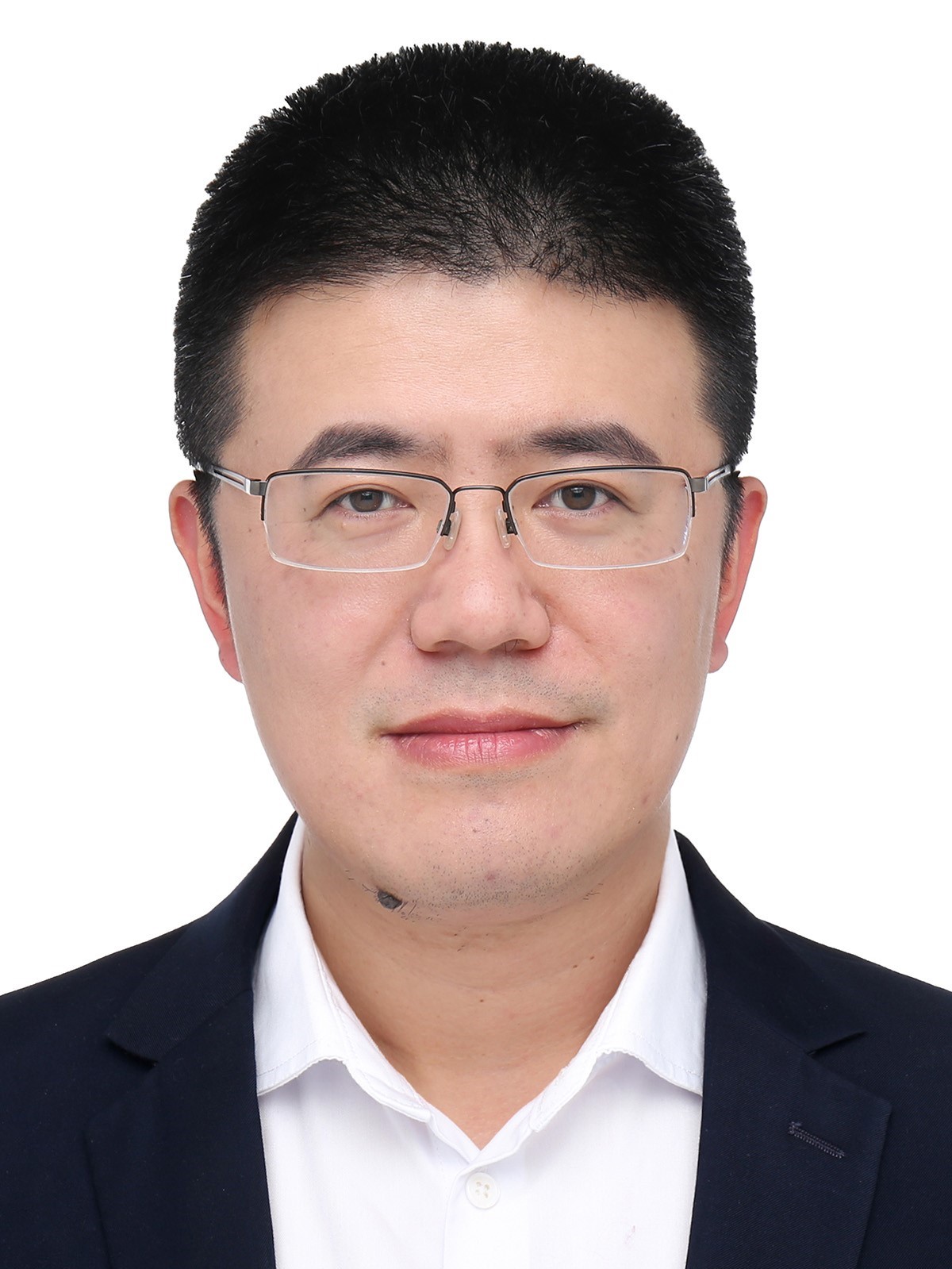}}]{Jiong Jin}
  
(M’11) received the B.E. degree with First Class Honours in Computer Engineering from Nanyang Technological University, Singapore, in 2006, and the Ph.D. degree in Electrical and Electronic Engineering from the University of Melbourne, Australia, in 2011. From 2011 to 2013, he was a Research Fellow in the Department of Electrical and Electronic Engineering at the University of Melbourne. He is currently an Associate Professor in the School of Software and Electrical Engineering, Swinburne University of Technology, Melbourne, Australia. His research interests include network design and optimization, edge computing and distributed systems, robotics and automation, and cyber-physical systems and Internet of Things as well as their applications in smart manufacturing, smart transportation and smart cities.
\end{IEEEbiography}

\begin{IEEEbiography}[{\includegraphics[width=1in,height=1.25in,clip,keepaspectratio]{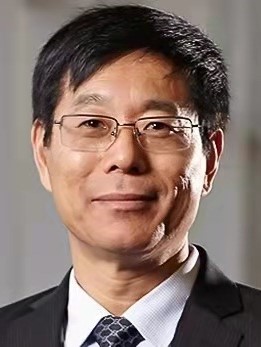}}]{Qing-Long~Han}
(M'09-SM'13-F'19) received the B.Sc. degree in Mathematics from Shandong Normal University, Jinan, China, in 1983,
and the M.Sc. and Ph.D. degrees in Control Engineering from East China University of Science and Technology, Shanghai, China, in 1992 and 1997, respectively.

Professor Han is Pro Vice-Chancellor (Research Quality) and a Distinguished Professor at Swinburne University of Technology, Melbourne, Australia. He held various academic and management positions at Griffith University and Central Queensland University, Australia. His research interests include networked control systems, multi-agent systems, time-delay systems, smart grids, unmanned surface vehicles, and neural networks.

Professor Han is a Highly Cited Researcher according to Clarivate Analytics. He is a Fellow of The Institution of Engineers Australia. He was one of Australia's Top 5 Lifetime Achievers (Research Superstars) in Engineering and Computer Science (The Australian's 2020 Research Magazine). He was the recipient of The 2021 M. A. Sargent Medal, which is awarded by Engineers Australia for longstanding eminence in science or the practice of electrical engineering. The prestigious M. A. Sargent Medal is the Highest Award of the Electrical College Board of Engineers Australia and consists of a Bronze Medal and a Certificate. He was the recipient of The 2020 IEEE Systems, Man, and Cybernetics (SMC) Society Andrew P. Sage Best Transactions Paper Award, The 2020 IEEE Transactions on Industrial Informatics Outstanding Paper Award, and The 2019 IEEE SMC Society Andrew P. Sage Best Transactions Paper Award. 

Professor Han is Co-Editor of Australian Journal of Electrical and Electronic Engineering, an Associate Editor for 12 international journals, including the IEEE TRANSACTIONS ON CYBERNETICS, the IEEE TRANSACTIONS ON INDUSTRIAL INFORMATICS, IEEE INDUSTRIAL ELECTRONICS MAGAZINE, the IEEE/CAA JOURNAL OF AUTOMATICA SINICA, Control Engineering Practice, and Information Sciences, and a Guest Editor for 13 Special Issues.
\end{IEEEbiography}
\end{document}